# DETEKSI DEPRESI DAN KECEMASAN PENGGUNA TWITTER MENGGUNAKAN BIDIRECTIONAL LSTM

**Kuncahyo Setyo Nugroho[1*)], Ismail Akbar[2)], Affi Nizar Suksmawati[3)], Istiadi[4)]**

[1)] Fakultas Ilmu Komputer, Universitas Brawijaya, Malang
[2)] Fakultas Sains dan Teknologi, UIN Maulana Malik Ibrahim, Malang
[3)] Fakultas Matematika dan Ilmu Pengetahuan Alam, Universitas Gadjah Mada, Yogyakarta
[4)] Fakultas Teknik, Universitas Widyagama Malang, Malang
**\*Email Korespondensi**: ksnugroho26@gmail.com

**ABSTRAK**

Gangguan mental yang paling umum dialami seseorang dalam kehidupan sehari-hari adalah depresi dan kecemasan. Stigma sosial membuat penderita depresi dan kecemasan diabaikan lingkungan sekitarnya. Oleh karena itu, mereka beralih ke media sosial seperti Twitter untuk mencari dukungan. Mendeteksi pengguna dengan potensi gangguan depresi dan kecemasan melalui data tekstual tidaklah mudah karena mereka tidak secara eksplisit berbicara tentang kondisi mentalnya. Dibutuhkan pemodelan yang mampu mengenali potensi pengguna yang mengalami depresi dan kecemasan pada data tekstual sehingga mereka mendapatkan penanganan lebih awal. Hal ini dapat dicapai dengan teknik klasifikasi teks. Salah satu pendekatan yang dapat digunakan adalah LSTM sebagai pengembangan aristektur RNN dalam menangani masalah *vanishing gradient*. LSTM standar tidak cukup menangkap informasi karena hanya mampu membaca kalimat dari satu arah. Sedangkan *Bidirectional* LSTM (BiLSTM) merupakan LSTM dua arah yang mampu menangkap informasi tanpa mengabaikan konteks dan arti dari suatu kalimat. Model BiLSTM yang diusulkan menunjukkan performa yang lebih tinggi daripada semua model *machine learning* tradisional dan LSTM standar. Berdasarkan hasil pengujian, akurasi tertinggi yang diperoleh BiLSTM mencapai 94.12%. Penelitian ini telah berhasil mengembangkan model untuk deteksi depresi dan kecemasan pengguna twitter.

**Kata kunci:** depresi dan kecemasan**,** *deep learning*, RNN, BiLSTM

***ABSTRACT***

*The most common mental disorders experienced by a person in daily life are depression and anxiety. Social stigma makes people with depression and anxiety neglected by their surroundings. Therefore, they turn to social media like Twitter for support. Detecting users with potential depression and anxiety disorders through textual data is not easy because they do not explicitly discuss their mental state. It takes a model that can identify potential users who experience depression and anxiety on textual data to get treatment earlier. Text classification techniques can achieve this. One approach that can be used is LSTM as an RNN architecture development in dealing with vanishing gradient problems. Standard LSTM does not capture enough information because it can only read sentences from one direction. Meanwhile, Bidirectional LSTM (BiLSTM) is a two-way LSTM that can capture information without ignoring the context and meaning of a sentence. The proposed BiLSTM model is higher than all traditional machine learning models and standard LSTMs. Based on the test results, the highest accuracy obtained by BiLSTM reached 94.12%. This study has succeeded in developing a model for the detection of depression and anxiety in Twitter users.*

## PENDAHULUAN

Gangguan mental didefinisikan sebagai sindrom yang secara klinis ditandai dengan regulasi emosi atau perilaku yang mencerminkan disfungsi dalam proses psikologis, biologis, atau perkembangan yang mendasari fungsi mental [1]. Gangguan mental menyebabkan penderitaan yang dapat menghambat aktivitas seseorang [2]. Terlepas dari dampak gangguan mental, adanya stigma sosial tentang gangguan mental merupakan penyakit jiwa yang tidak dapat disembuhkan membuat penderita diabaikan oleh lingkungan disekitarnya dan menghindari menjalani pengobatan yang diperlukan [3]. Gangguan mental yang paling umum adalah depresi dan kecemasan [4]. Diagnosis awal dan pengobatan merupakan hal penting yang harus dilakukan tepat waktu [5]. Namun, bagi penderita depresi dan kecemasan dibutuhkan keberanian dan kekuatan besar untuk mencari pengobatan yang tepat. Disisi lain, stigma gangguan mental membuat penderita depresi dan kecemasan beralih pada sumber daya *online* seperti media sosial Twitter untuk mencari dukungan [6]. Oleh karena itu, dibutuhkan suatu pemodelan yang mampu secara otomatis mengenali potensi seseorang mengalami depresi dan kecemasan sehingga memungkinkan diagnosis dan pengobatan yang tepat untuk penanganan lebih awal [7].

Deteksi depresi dan kecemasan melalui data tekstual telah dilakukan menggunakan *Support Vector Machine* (SVM) yang dibandingkan dengan *Bidirectional Encoder Representations from Transformers* (BERT) dan *A Lite* BERT (ALBERT) [8]. Performa model tertinggi diperoleh BERT dengan akurasi mencapai 75%. Penelitian lain menggunakan Naïve Bayes (NB) dan *Support Vector Regression* (SVR) [9]. Hasil pengujian pada 3.754 *tweet* menunjukkan SVR memperoleh akurasi lebih baik daripada NB sebesar 79.7%. Hasil pengujian juga dibandingkan dengan *K-Means Clustering* dan SVM. SVM memperoleh akurasi sebesar 78.8%, di mana SVM lebih baik dari NB tetapi masih dibawah SVR. Penelitian serupa pada klasifikasi teks telah dilakukan menggunakan *Bidirectional* LSTM (BiLSTM) [10]. Hasil pengujian dibandingkan dengan RNN, CNN, LSTM, dan NB menunjukkan *precision*, *recall*, dan F1-*score* tertinggi diperoleh BiLSTM. Meskipun RNN efektif mengekstrak informasi semantik antar kata, tetapi RNN tidak bisa menangani masalah hilangnya gradien pada kalimat panjang. Sedangkan *Long Short-Term Memory* (LSTM) dapat mengatasi masalah hilangnya gradien tetapi hanya sampai batas tertentu dengan membaca informasi satu arah. Oleh karena itu, BiLSTM diusulkan untuk mengatasi hilangnya gradien dengan mempertimbangkan membaca informasi dari dua arah [11].

Penelitian tentang deteksi depresi dan kecemasan pengguna Twitter pada bahasa Indonesia belum pernah dilakukan sebelumnya. Oleh karena itu, penelitian ini bertujuan melakukan prediksi depresi dan kecemasan pada data tekstual menggunakan BiLSTM. BiLSTM diusulkan karena mampu mengekstrak informasi kontekstual lebih cepat dengan pendekatan dua arah, sehingga tidak menghilangkan arti dan konteks suatu kalimat. Untuk mengevaluasi kinerja model, BiLSTM dibandingkan dengan beberapa metode *machine learning* tradisional lainnya seperti *k-Nearest Neighbor* (k-NN), *Support Vector Machine*(SVM), *Decision Tree Classifier* (DT), *Naïve Bayes* (NB) dan *Multi Layer Perceptron* (MLP). Selain itu, arsitektur LSTM umum juga dibandingkan dengan metode yang diusulkan.

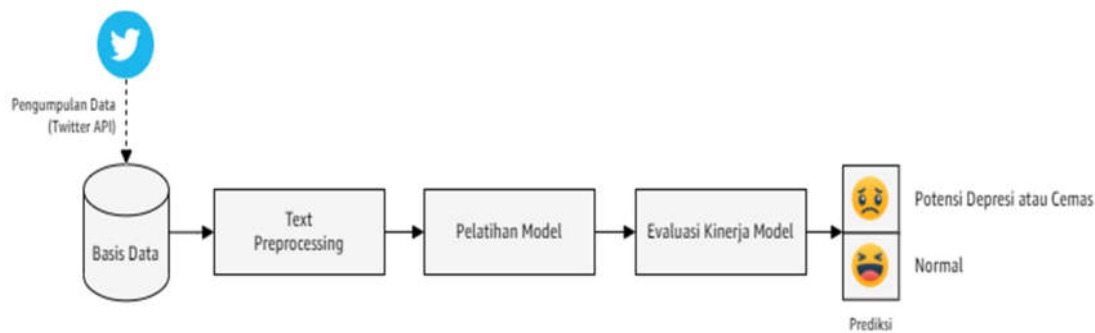

Gambar 1. Kerangka penelitian

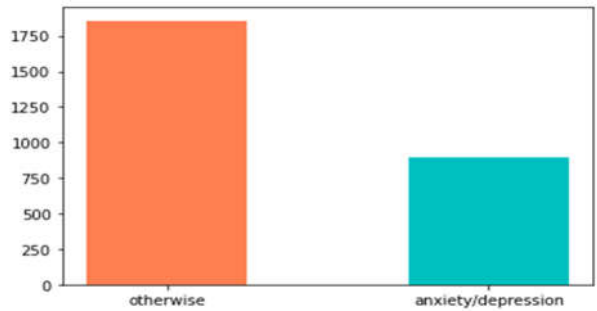

Gambar 2. Sebaran *tweet* berdasarkan label pada dataset

Tabel 1. Sampel data pada dateset

| Index | *Tweet* | Label |
|---|---|---|
| 5 | ngga enak bgt akhir2 ini rasanya, sering cemas berlebihan | 1 |
| 126 | Gak tau kenapa perasaan aku sedih gelisah y | 1 |
| 273 | Sedikit cemas banyak rindunya.... | 0 |
| 1789 | dulu dipaksa untuk menjadi yang paling cemas, sekarang terpaksa untuk jadi yang paling ikhlas 😊 | 0 |

## METODE PENELITIAN

Penelitian ini terdiri dari empat langkah utama yaitu pengumpulan dataset, *text-preprocessing*, pelatihan model, dan evaluasi kinerja model, seperti yang ditunjukkan kerangka penelitian pada Gambar 1.

### Dataset

Penelitian ini menggunakan dataset yang diperoleh dari media sosial Twitter dan telah dianotasi sebelumnya [12]. Dataset memiliki 2.751 *tweet* berbahasa Indonesia yang telah dikategorikan ke dalam dua label berbeda. Label 1 menyiratkan jika *tweet* pengguna memiliki potensi kecemasan, kegelisahan atau depresi, sedangkan label 0 adalah sebaliknya. Label 0 terdiri dari 1.857 *tweet* dan label 1 terdiri dari 894 *tweet*. Dataset memiliki distribusi kelas yang tidak seimbang seperti yang ditunjukkan pada Gambar 2, di mana label 0 memiliki jumlah *tweet* lebih banyak daripada label 1. Sampel data untuk setiap label ditunjukkan pada Tabel 1.

### Bidirectional LSTM

*Long Short-Term Memory* (LSTM) [13] adalah pengembangan arsitektur *Recurrent Neural Network* (RNN) [14] untuk menangani masalah *vanishing gradient*, di mana kemiringan fungsi kerugian menurun secara eksponensial pada saat memproses data sekuensial yang panjang [15]. Masalah ini menyebabkan RNN gagal menangkap *long term dependencies* [16] sehingga dapat mengurangi performa prediksi [17]. LSTM mengganti lapisan RNN dengan blok *memory cell* menggunakan mekanisme gerbang yang terdiri dari *forget gate, input gate*, dan *output gate* [11]. Sama halnya dengan RNN, LSTM tersusun atas *neuron* yang diproses secara berulang. Struktur *neuron* tunggal pada LSTM ditunjukkan pada Gambar 3.

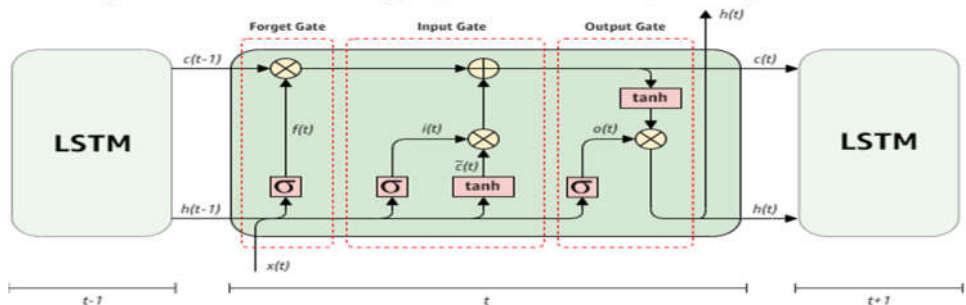

Gambar 3. Neuron tunggal pada arsitektur LSTM

*Forget gate* merupakan gerbang pertama pada LSTM untuk menentukan informasi mana yang akan dipertahankan atau dibuang dari *cell state*. Gerbang ini menerima *input* $h_{t-1}$ dan $X_t$ untuk menghasilkan nilai 0 atau 1 pada $C_{t-1}$ seperti diuraikan pada persamaan (1). Ketika *forget gate* bernilai 1, maka *cell state* akan menyimpan informasi, sedangkan jika bernilai 0 maka informasi akan dibuang dari *cell state*. Meningkatkan bias $b_f$ pada *forget gate* dapat meningkatkan kinerja LSTM [18].

$$f_t = \sigma(W_f \cdot [h_{t-1}, X_t] + b_f) \quad (1)$$

*Input gate* merupakan gerbang kedua pada LSTM untuk menentukan informasi apa yang akan disimpan pada *cell state*. Gerbang ini terdiri dari lapisan *sigmoid* dan lapisan *tanh*. Lapisan *sigmoid* memutuskan nilai mana yang akan diperbarui seperti diuraikan pada persamaan (2). Lapisan *tanh* membuat nilai baru $\tilde{C}_t$ untuk ditambahkan ke *cell state* seperti diuraikan pada persamaan (3). *Output* kedua lapisan ini digabungkan untuk memperbarui informasi *cell state*.

$$i_t = \sigma(W_i \cdot [h_{t-1}, X_t] + b_i) \quad (2)$$

$$\tilde{C}_t = tanh(W_c \cdot [h_{t-1}, X_t] + b_c) \quad (3)$$

Langkah berikutnya adalah memperbarui nilai *cell state* lama $C_{t-1}$ menjadi $C_t$ dengan mengalikan *cell state* lama dengan $f_t$ untuk menghapus nilai pada *forget gate* sebelumnya. Selanjutnya, ditambahkan dengan $i_t \tilde{C}_t$ sebagai nilai baru dan digunakan untuk memperbarui nilai *cell state* seperti diuraikan pada persamaan (4).

$$C_t = f_t \cdot C_{t-1} + i_t \cdot \tilde{C}_t \quad (4)$$

*Output gate* merupakan gerbang terakhir pada LSTM untuk menentukan *output* dari *cell state*. Pertama, lapisan *sigmoid* menentukan bagian dari *cell state* mana yang menjadi *output* seperti diuraikan pada persamaan (5). Selanjutnya, *output* tersebut dimasukkan kedalam lapisan *tanh* dan dikalikan dengan lapisan *sigmoid* agar *output* sesuai dengan yang diputuskan sebelumnya seperti diuraikan pada persamaan (6).

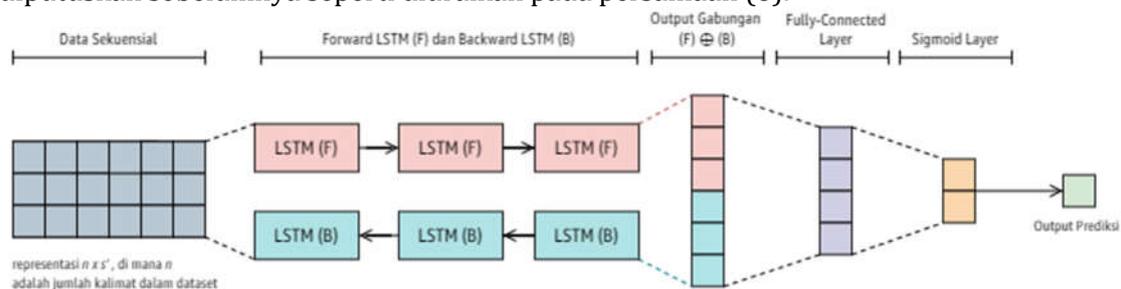

Gambar 4. Arsitektur BiLSTM yang diusulkan

$$O_t = \sigma(W_o \cdot [h_{t-1}, X_t] + b_o) \quad (5)$$

$$h_t = O_t \cdot \tanh(C_t) \quad (6)$$

Salah satu kelemahan LSTM adalah tidak cukup memperhitungkan informasi dari kata terakhir karena hanya membaca kalimat dari satu arah saja, awal ke akhir [19]. Oleh karena itu, kami menggunakan *bidirectional* LSTM (BiLSTM) untuk membaca kalimat dari dua arah sekaligus, awal ke akhir serta akhir ke awal. Secara teknis, BiLSTM menerapkan dua LSTM terpisah, satu untuk arah depan dan satu untuk arah mundur. Dua *hidden state* $h_t^{forward}$ dan $h_t^{backward}$ dari LSTM digabungkan menjadi *final hidden state* $h_t^{BiLSTM}$ seperti yang diuraikan pada persamaan (7). Sehingga, arsitektur BiLSTM yang kami usulkan disajikan pada Gambar 4.

$$h_t^{BiLSTM} = h_t^{forward} \oplus h_t^{backward} \tag{7}$$

**Evaluasi Kinerja Model**

*Confusion matrix* dapat digunakan untuk mengetahui kinerja model dengan menghitung rasio prediksi benar maupun salah, serta mengetahui jenis kesalahannya. *True positive* (TP) adalah kelas positif yang diprediksi benar. Misalnya, pengguna yang memiliki potensi kecemasan diprediksi memiliki kecemasan. *True negative* (TN) adalah kelas negatif yang diprediksi benar. Misalnya, pengguna yang tidak memiliki potensi kecemasan diprediksi tidak memiliki kecemasan. *False positive* (FP) adalah kelas negatif yang diprediksi sebagai kelas positif. Misalnya, pengguna yang tidak memiliki kecemasan diprediksi memiliki potensi kecemasan. *False negative* (FN) adalah kelas positif yang diprediksi sebagai kelas negatif. Misalnya, pengguna yang memiliki kecemasan diprediksi tidak memiliki potensi kecemasan.

Metrik yang paling sering digunakan untuk mengevaluasi model berdasarkan *confusion matrix* adalah *accuracy*. *Accuracy* adalah rasio prediksi benar (TP dan TN) dengan keseluruhan data yang menggambarkan tingkat kedekatan nilai prediksi dengan nilai sebenarnya, seperti diuraikan pada persamaan (8). Permasalahan distribusi data tidak seimbang adalah sampel data negatif lebih banyak daripada data positif. Oleh karena itu, kami menggunakan dua metrik lainnya, *precision* dan *recall*. *Precision* adalah rasio prediksi benar positif (TP) dengan keseluruhan data yang diprediksi positif, seperti diuraikan pada persamaan (9). Sedangkan, *recall* adalah rasio prediksi benar positif (TP) dengan keseluruhan data yang benar positif, seperti diuraikan pada persamaan (10).

$$accuracy = \frac{TP + TN}{TP + FP + FN + TN} \tag{8}$$

$$precision = \frac{TP}{TP + FP} \tag{9}$$

$$recall = \frac{TP}{TP + FN} \tag{10}$$

**HASIL DAN PEMBAHASAN**

Dataset yang tersedia memiliki format tidak terstruktur. Oleh karena itu, langkah pertama yang dilakukan adalah *text preprocessing* meliputi menghapus angka, menghapus URL, menghapus *username mention*, dan menghapus tanda baca. Sedangkan *stemming*, *stopword removal* dan normalisasi kata slang atau kata gaul tidak dilakukan karena kami tidak ingin mengubah arti dan konteks dari suatu kalimat. Seluruh eksperimen dilakukan pada Google Colab[1] *environment* menggunakan Python 3.6 dengan spesifikasi 1 Tesla V100-SXM2-16 GB GPU dan 27.8 GB RAM. Dataset dibagi menjadi tiga bagian, yaitu data latih, data uji dan data validasi. Pertama, dataset utuh dibagi 80% untuk data latih, sisanya untuk data uji. Kemudian, data latih dibagi dua untuk data validasi selama proses pelatihan model.

Sebagai *baseline* model, kami menggunakan beberapa metode *machine learning* tradisional meliputi *k-Nearest Neighbor* (k-NN), *Support Vector Machine* (SVM), *Decision Tree Classifier* (DT), *Naïve Bayes* (NB) dan *Multi Layer Perceptron* (MLP). Nilai parameter yang telah ditentukan untuk setiap *baseline* model disajikan pada Tabel 2. Kami menggunakan skema pembobotan kata *Term Frequency -Inverse Document Frequency* (TF-IDF) dengan kombinasi *bi-gram* sebagai metode ekstraksi fitur. Selanjutnya, prosedur validasi silang 10 lipat diterapkan pada data latih selama fase pelatihan model. Hasil pengujian *baseline* model disajikan pada Tabel 3.

---

[1] https://colab.research.google.com

Berdasarkan Tabel 3 dapat diketahui bahwa MLP memiliki akurasi tertinggi pada fase pelatihan sebesar 0.9850 maupun fase pengujian sebesar 0.7422 daripada model lainnya. Akurasi validasi silang tertinggi juga diperoleh oleh MLP sebesar 0.76 dengan standar deviasi $\pm$ 0.0628 . Sedangkan akurasi pengujian terendah adalah 0.6497 diperoleh DT, meskipun memiliki akurasi pelatihan yang sama dengan MLP. Jika diperhatikan, akurasi pelatihan dan akurasi pengujian memiliki selisih yang jauh. Hal ini dapat disebabkan karena model terlalu naif. Hasil berbeda diperoleh akurasi validasi silang dengan memiliki nilai yang cenderung dekat dengan akurasi pengujian.

Tabel 2. Nilai parameter untuk *baseline* model

| *Baseline* Model | Nama dan Nilai Parameter |
| --- | --- |
| k-NN | n_neighbors=3 |
| SVM | kernel= polynomial, C=1.0, degree=3 |
| DT | criterion=gini, min_samples_split=2, min_samples_leaf=1 |
| NB | alpha=1.0 |
| MLP | hidden_layer_size=25, solver=adam, learning_rate=1e-3, max_iter=100 |

Tabel 3. Hasil pengujian *baseline* model

| *Baseline* Model | Akurasi Pelatihan | Akurasi Validasi Silang | Akurasi Pengujian |
| --- | --- | --- | --- |
| k-NN | 0.7995 | 0.6786 ($\pm$ 0.0645) | 0.6588 |
| SVM | 0.9831 | 0.6836 ($\pm$ 0.0688) | 0.6696 |
| DT | 0.9850 | 0.7263 ($\pm$0.0322) | 0.6497 |
| NB | 0.8945 | 0.7286 ($\pm$0.0633) | 0.6987 |
| MLP | 0.9850 | 0.7600 ($\pm$ 0.0628) | 0.7422 |

Pengujian selanjutnya adalah menerapkan arsitektur BiLSTM seperti yang ditunjukkan pada Gambar 4. Dataset yang telah melalui *text-preprocessing* kemudian dipisahkan berdasarkan spasi menggunakan *tokenizer* dari *library* Keras[2]. Berikutnya, daftar token kosakata dikonversi menjadi urutan numerik dengan mengganti indeks setiap kosakata dengan nilai *integer*. Setiap token kata dipetakan ke vektor berukuran $s$, di mana $s$ adalah jumlah kata dalam sebuah kalimat. Kami menerapkan strategi *zero-padding* sehingga semua kalimat memiliki dimensi vektor yang sama $X \in R^{s'}$ dengan nilai $s' = 1000$. Sebagai pembanding, kami juga menerapkan arsitektur LSTM yang umum. Parameter yang telah ditentukan untuk LSTM maupun BiLSTM disajikan pada Tabel 4. Jumlah *epoch* ditentukan sebanyak 25 kali untuk setiap percobaan. Sedangkan untuk menghindari *over-fitting* pada model selama fase pelatihan, kami menentukan nilai *dropout* sebesar 0.5.

Tabel 4. Pengaturan parameter LSTM dan BiLSTM

| Nama Parameter | Nilai Parameter |
| --- | --- |
| Embedding_size | 200 |
| activation | sigmoid |
| optimizer | adam |
| learning_rate | 1e-3 |
| batch_size | 64 |
| regularizer | L2 |

---

[2] https://keras.io/api/preprocessing/text

Tabel 5. Hasil pengujian model LSTM dan BiLSTM

| Model | Akurasi | *Training Loss* | *Precision* | *Recall* |
|---|---|---|---|---|
| LSTM | 0.8491 | 0.3707 | 0.7659 | 0.7673 |
| BiLSTM | 0.9412 | 0.1826 | 0.9759 | 0.8386 |

Hasil pengujian LSTM dan BiLSTM pada Tabel 5 menunjukkan BiLSTM memiliki kinerja yang lebih baik pada semua metrik evaluasi. BiLSTM juga unggul jika dibandingkan dengan semua model *machine learning* tradisional pada Tabel 3. Akurasi pengujian tertinggi adalah 0.9412 dengan *training loss* sebesar 0.1826. Sedangkan *precision* dan *recall* yang diperoleh adalah 0.9759 dan 0.8386. Berdasarkan grafik fase pelatihan pada **Error! Reference source not found.** dan Gambar 6, dapat diketahui bahwa BiLSTM memiliki akurasi pelatihan dan *training loss* yang lebih stabil pada setiap *epoch*nya. Sedangkan LSTM pada awal *epoch* cenderung memiliki akurasi yang kecil tetapi mengalami peningkatan pada setiap *epoch* berikutnya. Sama halnya dengan *training loss* yang menunjukkan penurunan pada setiap *epoch*nya yang berarti model telah belajar selama fase pelatihan.

Arsitektur BiLSTM yang diusulkan dalam penelitian ini menunjukkan peningkatan kinerja dibandingkan dengan *baseline* model maupun LSTM standar. Hal ini juga memperkuat fakta bahwa pendekatan *deep learning* mampu mencapai performa yang lebih baik daripada pendekatan *machine learning* tradisional. Kami juga mengamati bahwa BiLSTM mampu mengatasi masalah *long-term dependency*. Keuntungan lain dari penelitian ini adalah kemampuan BiLSTM untuk membaca informasi dari dua arah sekaligus.

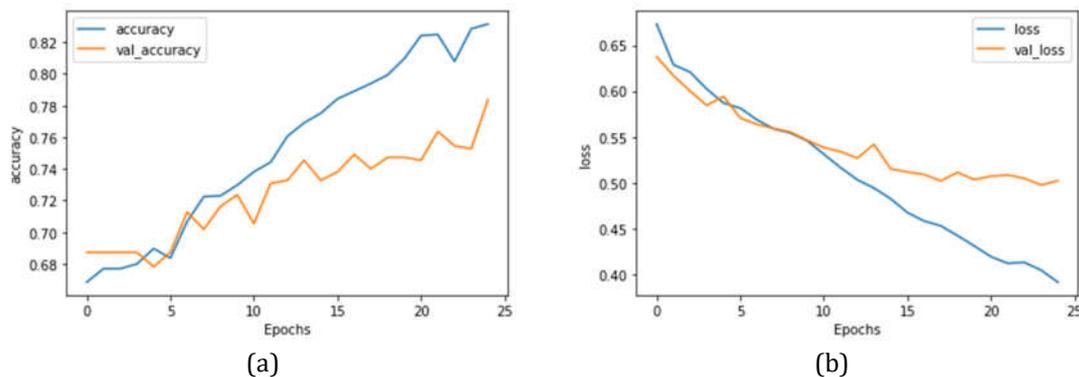

(a)      (b)

Gambar 5. Grafik pada fase pelatihan LSTM, (a) akurasi dan validasi pelatihan, (b) training *dan* validation loss

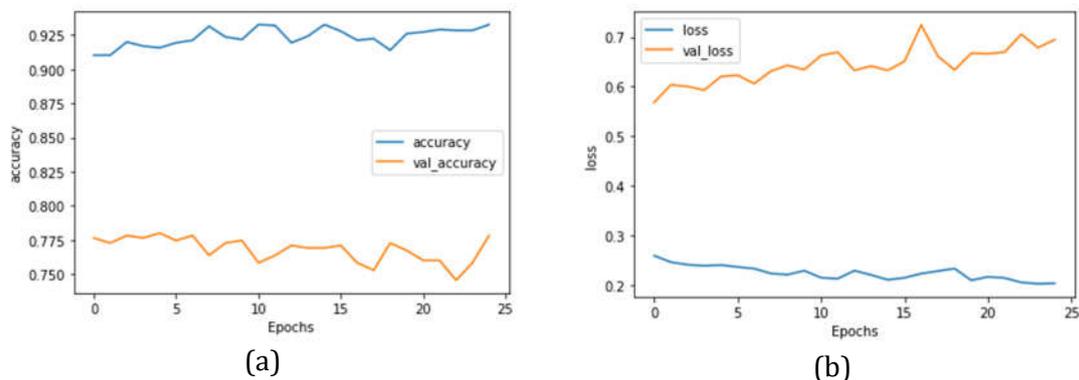

(a)      (b)

Gambar 6. Grafik pada fase pelatihan BiLSTM, (a) akurasi dan validasi pelatihan, (b) training dan validation loss

Sedangkan kelemahan LSTM maupun BiLSTM adalah membutuhkan lebih banyak data serta waktu dan biaya komputasi yang lebih tinggi daripada *baseline* model yang ada. Dengan demikian, kemampuan BiLSTM dalam membaca konteks melalui dua arah sekaligus memberikan hasil yang baik pada deteksi depresi dan kecemasan pengguna Twitter.

**KESIMPULAN**

Pada penelitian ini kami mengusulkan arsitektur BiLSTM untuk deteksi depresi dan kecemasan pengguna Twitter pada bahasa Indonesia. Berdasarkan hasil pengujian, model kami menunjukkan performa yang lebih tinggi daripada semua model *machine learning* tradisional dan LSTM standar. Akurasi tertinggi yang diperoleh menggunakan BiLSTM mencapai 94.12%. Hal ini dapat diraih karena BiLSTM mampu mengambil informasi dengan membaca konteks melalui dua arah sekaligus. Namun, BiLSTM membutuhkan dataset yang cukup besar untuk menghindari model yang *over-fitting*. Selain itu, biaya dan waktu komputasi yang dibutuhkan juga tinggi. Pada penelitian berikutnya, kombinasi *word embedding* perlu diterapkan agar menghasilkan representasi kata yang lebih kaya. Selain itu, *hyperparameter-tuning* juga perlu dilakukan guna meningkatkan performa model.

**REFERENSI**